\documentclass[conference]{IEEEtran}
\IEEEoverridecommandlockouts

\def\BibTeX{{\rm B\kern-.05em{\sc i\kern-.025em b}\kern-.08em
    T\kern-.1667em\lower.7ex\hbox{E}\kern-.125emX}}

\title{MUSDA: Multi-source Multi-modality Unsupervised Domain Adaptive 3D Object Detection for Autonomous Driving
}

\author{
\IEEEauthorblockN{Xiaohu Lu}
\IEEEauthorblockA{\textit{Electrical and Computer Engineering} \\
\textit{Michigan State University}\\
East Lansing, USA \\
luxiaohu@msu.edu}
\and
\IEEEauthorblockN{Hamed Khatounabadi}
\IEEEauthorblockA{\textit{Electrical and Computer Engineering} \\
\textit{Michigan State University}\\
East Lansing, USA \\
khatouna@msu.edu}
\and
\IEEEauthorblockN{Hayder Radha}
\IEEEauthorblockA{\textit{Electrical and Computer Engineering} \\
\textit{Michigan State University}\\
East Lansing, USA \\
radha@msu.edu}
}

\usepackage{amsmath}
\usepackage{bm}
\usepackage{multirow}
\usepackage{color}
\usepackage{makecell}
\usepackage{epsfig}
\usepackage{graphicx}
\usepackage{amssymb}

\newcommand{\equ}{Eq.}
\newcommand{\tab}{Tab.}
\newcommand{\fig}{Fig.}

\newcommand{\ssec}{Sec.}

\usepackage[capitalize]{cleveref}
\crefname{section}{Sec.}{Secs.}
\Crefname{section}{Section}{Sections}
\Crefname{table}{Table}{Tables}
\crefname{table}{Tab.}{Tabs.}

\definecolor{greenmine}{rgb}{0.85, 0.85, 0.85}
\begin{document}
\maketitle
\begin{abstract}

With the advancement of autonomous driving, numerous annotated multi-modality datasets have become available. This presents an opportunity to develop domain-adaptive 3D object detectors for new environments without relying on labor-intensive manual annotations. However, traditional domain adaptation methods typically focus on a single source domain or a single modality, limiting their effectiveness in multi-source, multi-modality scenarios. In this paper, we propose a novel framework for multi-source, multi-modality unsupervised domain adaptation in 3D object detection for autonomous driving. Given multiple labeled source domains and one unlabeled target domain, our framework first introduces \textbf{hierarchical spatially-conditioned (HSC) domain classifiers}, which jointly align features from both camera and LiDAR modalities at two distinct levels for each source–target domain pair. To effectively leverage information from multiple source domains, we construct a prototype graph between each pair of domains. Based on this, we develop a \textbf{prototype graph weighted (PGW) multi-source fusion} strategy to aggregate predictions from multiple source detection heads. Experimental results on three widely used 3D object detection datasets — Waymo, nuScenes, and Lyft — demonstrate that our proposed framework effectively integrates information across both modalities and source domains, consistently outperforming state-of-the-art methods.

\end{abstract}    
\section{Introduction}

With increase deployment of self-driving cars, the perception systems of autonomous vehicles must now operate reliably across a wide range of scenarios, including adverse weather, varying lighting conditions, and other challenging environments. These requirements necessitate the integration of multiple sensors to enhance perception performance and ensure safety. Consequently, an increasing number of datasets have been released with multimodal data. Early examples include KITTI~\cite{geiger2012we}, Lyft Level 5~\cite{lyft2019}, nuScenes~\cite{caesar2020nuscenes}, Waymo~\cite{sun2020scalability}, and DENSE~\cite{bijelic2020seeing}, followed by more recent datasets such as Ithaca365~\cite{diaz2022ithaca365}, Argoverse 2~\cite{wilson2023argoverse}, MSU-Four Seasons~\cite{kent2024msu}, and MUSES~\cite{brodermann2025muses}, among others. With dozens of annotated multi-modal datasets now available for autonomous driving, it becomes increasingly valuable to explore how the industry and academia can leverage these resources to develop domain-adaptive 3D object detectors for new environments in the absence of labor-intensive manual annotations.

Traditional domain adaptation, involving a single source domain and a single modality, has been extensively studied in tasks such as image classification~\cite{peng2024unsupervised,nejjar2023dare}, semantic segmentation~\cite{colomer2023adapt,cheng2023adpl}, and object detection~\cite{lu2024dali,kennerley20232pcnet,mattolin2023confmix}. These approaches aim to transfer knowledge from a single labeled source domain to improve performance on an unlabeled target domain within the same modality. However, extending from traditional domain adaptation to the multi-source, multi-modality setting introduces significantly greater challenges. These challenges arise from two key aspects: (1) how to effectively fuse domain-adaptive features across different modalities, and (2) how to efficiently exploit complementary information from multiple source domains. To tackle the multi-modality aspect in object detection, several methods have been proposed~\cite{hegde2024multimodal,chang2024cmda,tsai2022see,eskandar2022unsupervised,bijelic2020seeing}. Meanwhile, other approaches focus on the multi-source setting~\cite{zhao2024multi,belal2024multi,tsai2024ms3d++,tsai2023ms3d,wu2023towards,wu2022target,zhang2022multi,yao2021multi,lin2021domain}, aiming to leverage diverse domain knowledge to improve generalization. However, as of the time of writing of this manuscript and to the best of our knowledge, no existing work has addressed multi-source multi-modality domain-adaptive 3D object detection for autonomous driving.

Consequently, in this paper, we address this arguably new problem of multi-source multi-modality unsupervised domain adaptation (UDA) for 3D object detection in autonomous driving. Specifically, we propose hierarchical spatially-conditioned domain classifiers that first condition a multi-modality feature map using a predicted heatmap. The resulting \textit{multi-modality domain probability map} is then used as a conditioning signal for the modality-specific feature maps from LiDAR and camera inputs, enabling fine-grained alignment between source and target domain features. To effectively leverage the diverse information across different source domains, we introduce a low-complexity modification to the baseline architecture by incorporating a Bird’s Eye View (BEV) domain embedding, which learns domain-specific representations. During inference, the input LiDAR and camera data are processed by a shared backbone network and passed through the detection head multiple times, each time with a different BEV domain embedding to generate multiple predictions that correspond to the multi-source domain-specific representations. To guide the fusion of these multi-source predictions during inference, we further propose a novel prototype graph, resulting in improved detection performance. Our \textbf{MU}lti-\textbf{S}ource multi-modality unsupervised \textbf{D}omain \textbf{A}daptive 3D object detection for autonomous driving is referred to as MUSDA for the remainder of this paper. In summary, our contributions are:
    \begin{enumerate}
      \item We propose a novel hierarchical spatially-conditioned domain classifiers to align multi-modality domain features in both modality-fused feature map and modality-specific feature maps. 
      \item We introduce a novel BEV domain embedding strategy to learn domain-specific information in multi-source scenario.      
      \item We develop a prototype graph to model relationships between multiple domains and use it to weight and fuse predictions from different sources.
      \item Experiments on three popular 3D object detection datasets with both LiDAR and camera modalities, i.e.,  Waymo~\cite{sun2020scalability}, nuScenes~\cite{caesar2020nuscenes}, and Lyft~\cite{lyft2019} demonstrate the effectiveness of the proposed MUSDA framework.
    \end{enumerate}

    \begin{table}[!ht]
        \centering
        %\vspace{-0.3cm}
    	\caption{Time consumption for different components in the baseline method BEVFusion~\cite{liu2023bevfusion}.}
    	\label{tab:components_time_consumption}
    	\small
    	\setlength{\tabcolsep}{3.5mm}
        \renewcommand{\arraystretch}{1.1}
    	\begin{tabular}{l c c}
    	\hline
    	Name  &Time (ms) &Percentage \\
    	\hline
    	Camera Backbone    &55.8    &38.8\% \\
            LiDAR Backbone     &73.5    &51.2\% \\
            Fusion Backbone    &1.1     &0.8\% \\
            Fusion Head        &13.2    &9.2\% \\
    	\hline
        \end{tabular}
    \end{table}

\section{Methodology}

    \begin{figure*}
    	\centering
    	\footnotesize
    	\begin{tabular}{c}
    		\includegraphics[width=0.98\linewidth]{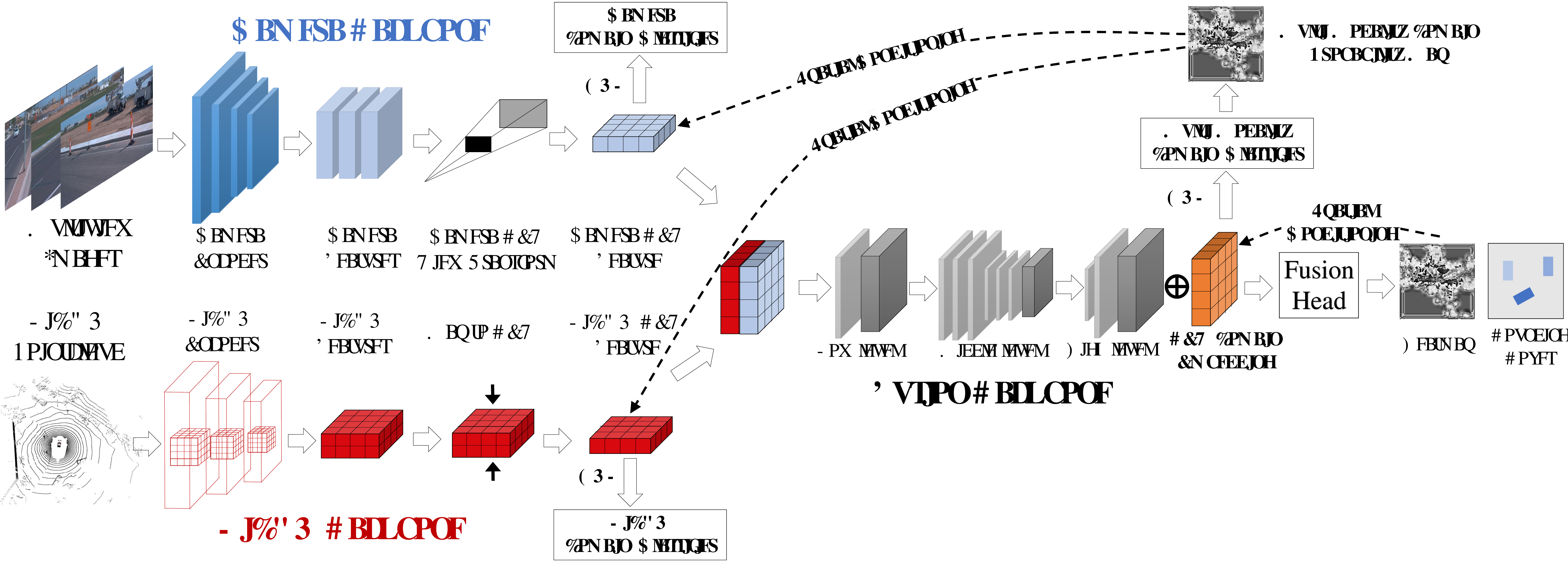}
    	\end{tabular} 
    	\caption{The framework of our single-source multi-modality unsupervised domain adaptive approach. We first use the predicted heatmap as a first-level spatial condition to guide the multi-modality domain classifier. Then, the predicted multi-modality domain probability map is employed as a second-level spatial condition to further guide the camera and LiDAR BEV features.}
    	\label{fig:framework}
    	%\vspace{-1.0em}
    \end{figure*}

\subsection{Hierarchical Spatially-conditioned (HSC) Domain Classifiers}
\label{sec:spatial_domain_classifier}
In unsupervised domain adaptation, domain classifiers~\cite{ganin2016domain} combined with gradient reversal layers (GRL) are widely used to encourage the backbone network to learn domain-invariant features that confuse the domain classifier. Several previous methods have extended this approach from image-level to pixel-level features~\cite{hsu2020every}, and from single-scale to multi-scale representations~\cite{hnewa2021multiscale}. To address the complex multi-modality scenario, where features originate from images, point clouds, and their fusion, we propose a novel hierarchical spatially-conditioned domain classifiers strategy that aligns features from both camera and LiDAR modalities across each source-target domain pair. Specifically, let $G_f^{2D}$, $G_f^{3D}$, and  $G_f^{MM}$ represent the camera backbone network, LiDAR backbone network, and multi-modality fusion backbone network, respectively. Given a point cloud $\mathcal{P}$ and the corresponding multi-view images $\mathcal{I}$, the camera backbone network generates camera BEV features $f^{2D} = G_f^{2D}(\mathcal{I})$, and the LiDAR backbone network generates LiDAR BEV features $f^{3D} = G_f^{3D}(\mathcal{P})$. These 2D and 3D BEV features are concatenated and fed into the fusion backbone to generate multi-modality features $f^{MM} = G_f^{MM}([f^{2D}, f^{3D}])$. Finally the fusion head $Y$ takes $f^{MM}$ and the BEV domain embedding $f^{emb}$ (explained in \ssec~\ref{sec:domain_embedding}) as input to generate the predictions of the bounding boxes and the heatmap $h$. 

To eliminate the influence of background pixels and enable more precise feature alignment on foreground regions, we use the predicted heatmap as a spatial condition to guide the multi-modality domain classifier operating on the multi-modality feature. Specifically, given the multi-modality feature $f^{MM}\in \mathbb{R}^{C\times H \times W}$ and the predicted heatmap $h\in \mathbb{R}^{K\times H \times W}$, where $K$ is the number of classes, we first calculate the class-agnostic heatmap $h^{'} \in \mathbb{R}^{H \times W}$ as the maximum probability along the class dimension of $h$. The multi-modality feature is then weighted by the class-agnostic heatmap as the first-level spatial condition and passed to the multi-modality domain classifier $D^{MM}$ to predict the multi-modality domain probability map as below:
\begin{equation}
\label{equ:domain_probability_mm}
p^{MM} = D^{MM}(GRL(f^{MM} \odot h^{'})) \in \mathbb{R}^{H \times W}.
\end{equation}
%Note that the multi-modality probability map $p^{MM}$ consists of two channels, $p_{1}^{MM}$ and $p_{2}^{MM}$, which represent the source and target domains' probability maps, and where $p_{1}^{MM}=1-p_{2}^{MM}$.
The corresponding domain classification loss is calculated pixel-wisely as:
\begin{equation} \label{Eq:domain_loss_mm}
    L_{\text{dom}}^{MM} = - \frac{1}{uv}\sum_{u, v} \left[ d \log p^{MM}_{u,v} + (1 - d) \log (1 - p^{MM}_{u,v}) \right],
\end{equation} 
where $u$ and $v$ denote spatial location on the feature map, \( d \in \{0, 1\}\) is the domain label, with $d=0$ for the target domain and $d=1$ for the source domains, and \( p_{u,v} \) represents the predicted domain probability at position \((u, v)\) for the multi-modality feature map. GRL denotes the gradient reversal layer, which inverts the gradient during backpropagation.

Aligning the fused multi-modality features may be insufficient in complex multi-modality UDA scenarios. Therefore, we further design domain classifiers for both the camera and LiDAR modalities using a hierarchical spatially-conditioned strategy. Specifically, we use the multi-modality domain probability map as the second-level spatial condition to the camera and LiDAR BEV feature, which is then passed to the corresponding single-modality domain classifier $D^{2D}$ and $D^{3D}$ to predict the single-modality domain probability map as below:
\begin{equation}
\label{equ:domain_probability_mm}
p^{SM} = D^{SM}(GRL(f^{SM} \odot p^{MM})) \in \mathbb{R}^{H \times W}.
\end{equation}
where $SM \in \{2D, 3D\}$ denotes the single modality (either 2D or 3D).
%, and $f^{SM} \otimes p^{MM} = [f^{SM} \odot p^{MM}_{1},\ f^{SM} \odot p^{MM}_{2}]$ represents the concatenation of the element-wise products between the single-modality feature $f^{SM}$ and the first and second channels of the multi-modality domain probability map $p^{MM}$. 
The same domain classification loss as in \equ~\ref{Eq:domain_loss_mm} is used to train the 2D and 3D domain classifiers. 

In summary, our hierarchical spatially-conditioned domain classifiers operate in two levels: the first-level domain classifier uses a class-agnostic heatmap as a spatial condition to separate foreground from background, while the second-level domain classifiers use the domain probability map from the first level to further distinguish features that are easy or hard for the domain classifier to confuse. This hierarchical design leads to improved performance, as demonstrated in the experimental results.

The overall training objective is defined as:
\begin{equation}
L = L_{det} + \lambda (L_{dom}^{MM} + L_{dom}^{3D} + L_{dom}^{2D}),
\end{equation}
where $\mathcal{L}_{\text{det}}$ is the detection loss computed on the heatmap and bounding boxes, following the formulation in~\cite{liu2023bevfusion}; $\mathcal{L}_{\text{dom}}^{MM}$, $\mathcal{L}_{\text{dom}}^{3D}$, and $\mathcal{L}_{\text{dom}}^{2D}$ denote the domain classification losses for the multi-modality, 3D, and 2D, respectively; and $\lambda$ is a weighting factor that balances the effect of domain classification.

\subsection{Prototype Graph Weighted (PGW) Multi-source Fusion}

In a multi-source unsupervised domain adaptation (UDA) object detection scenario, we consider $N$ labeled source domains, denoted as $S_1, S_2, \dots, S_N$, and a single unlabeled target domain $T$. A key challenge in this setting is that different source domains often exhibit varying domain gaps across object classes. For instance, as shown in \tab~\ref{tab:multi_dataset}, Lyft shares more similar car sizes with Waymo, while nuScenes offers significantly more pedestrian annotations than Lyft, reflecting distinct source-domain-specific characteristics. To address this phenomenon, we extend our single-source multi-modality framework (shown in \fig~\ref{fig:framework}) to a multi-source setting that is designed to handle variations in domain gaps between different source-target pairs. In that context, we develop a graph-based multi-source fusion strategy to effectively leverage source-domain-specific information, as illustrated in \fig~\ref{fig:multi_source_fusion}.

    \begin{table*}[!ht]
        \centering
        %\vspace{-0.3cm}
    	\caption{Annotations per frame and Mean size of three different object classes: Car, Pedestrian, Cyclist in Waymo, nuScenes, and Lyft.}
    	\label{tab:multi_dataset}
    	\small
    	\setlength{\tabcolsep}{3.0mm}
    	\begin{tabular}{l |c |c |c |c |c |c}
    	\hline
            \multirow{2}{*}{Dataset}  &\multicolumn{3}{c|}{$\#$Annotations Per Frame} &\multicolumn{3}{c}{Mean Size (m)} \\
            \cline{2-7}
            &Car &Pedestrian &Cyclist &Car &Pedestrian &Cyclist \\
    	\hline
    	Waymo     &27.5 &12.8 &0.3   &[4.81, 2.11, 1.78]    &[0.91, 0.86, 1.74] &[1.81, 0.84, 1.77]\\
            nuScenes  &18.9 &6.6 &0.6    &[5.43, 2.11, 2.04]    &[0.72, 0.66, 1.76] &[1.91, 0.69, 1.38]\\
            Lyft      &25.9 &1.0 &1.0    &[5.20, 2.01, 1.87]    &[0.80, 0.77, 1.78] &[1.77, 0.64, 1.45]\\
    	\hline
        \end{tabular}
    \end{table*}
    
\subsubsection{Multi-source Training With Domain Embedding}
\label{sec:domain_embedding}
\begin{figure*}
    	\centering
    	\footnotesize
    	\begin{tabular}{c}
    		\includegraphics[width=0.80\linewidth]{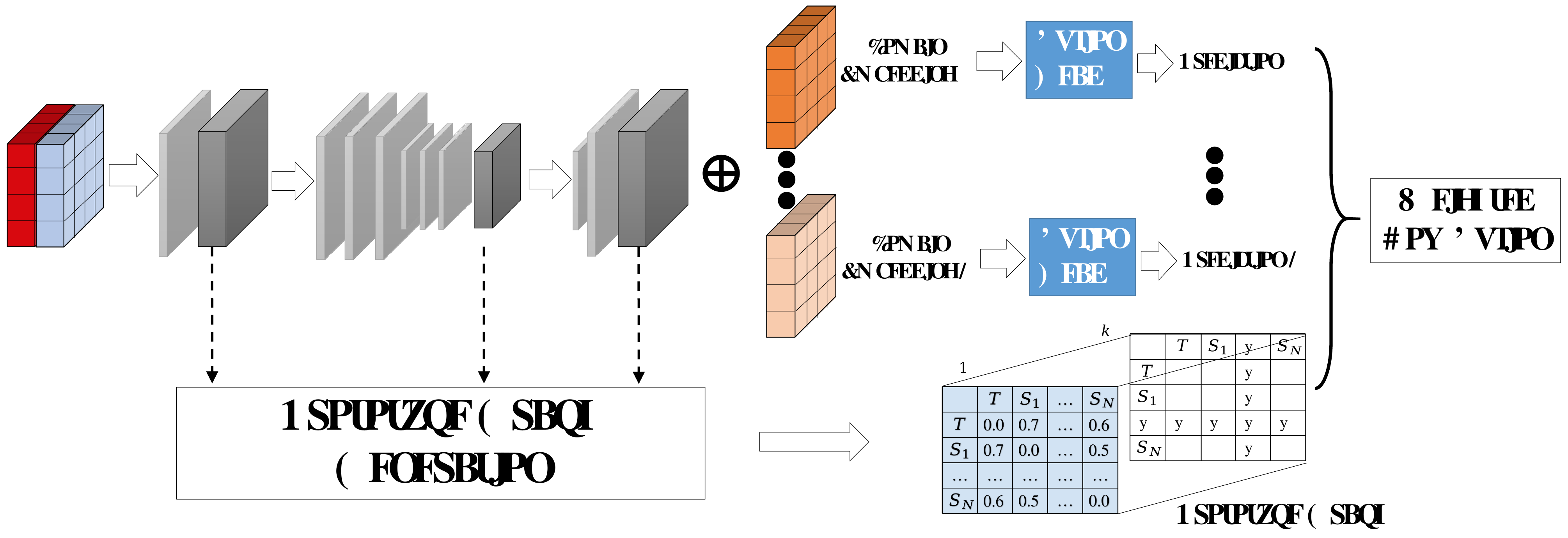}
    	\end{tabular} 
    	\caption{Overview of our prototype graph weighted (PGW) multi-source fusion approach. During training, a domain embedding is learned to capture source-domain-specific characteristics. At inference time, the multi-modality feature map is concatenated with different domain embeddings and passed through the detection head multiple times to generate diverse predictions. A prototype graph is then constructed and used to weight and fuse these predictions for improved detection performance.}
    	\label{fig:multi_source_fusion}
    	%\vspace{-1.0em}
    \end{figure*}

Training a unified model that can effectively learn from multiple source domains while minimizing the domain gap with the target domain is challenging, as different source domains may have varying levels of similarity to the target domain depending on the object class. A naive approach is to use separate fusion backbones for each source domain, but this significantly increases model complexity. Noting that these separate backbones largely share functionality and differ mainly in source-domain-specific information, we propose to train a single shared fusion backbone for all source domains. To capture source-domain-specific characteristics more efficiently, we introduce a domain embedding of size $N \times \text{dim}$, where $N$ is the number of source domains. The embedding dimension is set to $\text{dim}=128$ by default in all experiments. As shown in \fig~\ref{fig:multi_source_fusion}, during each training iteration involving the $i$-th source domain, the corresponding domain embedding vector is expanded from $\mathbb{R}^{1\times \text{dim}}$ to $\mathbb{R}^{\text{dim}\times H\times W}$ to match the spatial dimensions of the multi-modality feature $f^{MM}\in \mathbb{R}^{C\times H \times W}$ and then concatenated with it on the channel dimension. The resulting feature is then passed through the corresponding fusion head to compute the detection loss $Loss_{det}$. During training, the domain embedding is updated in the first half of the total epochs and frozen in the second half to enhance stability. Specifically, in the first half, the model learns representative embeddings for different domains, while in the second half, these embeddings are used to guide feature learning.

\subsubsection{Multi-source Fusion Via Prototype Graph}
To effectively capture domain-specific information across different source domains, we propose constructing a prototype graph to capture class-wise relationships between domains, and use this graph to guide the fusion of predictions from multiple sources. Specifically, after completing the training stage with all source domains and the target domain (with pseudo labels, as described in \ssec~\ref{sec:domain_embedding}), we input every training sample from each domain into the trained model to extract low-, middle-, and high-level multi-modality features along with the predicted heatmap. For the $i$-th feature map $f^{MM}_i \in \mathbb{R}^{C_i \times H_i \times W_i}$, where $i \in \{1,2,3\}$ corresponds to the low-, middle-, and high-level features, we interpolate the predicted heatmap $h \in \mathbb{R}^{K \times H \times W}$ to obtain $h_i \in \mathbb{R}^{K \times H_i \times W_i}$, which matches the spatial dimensions of $f^{MM}_i$. Then we flatten the spatial dimensions of the interpolated heat maps to get $f^{\text{flat}}_i \in \mathbb{R}^{C_i \times M_i}$ and $h^{\text{flat}}_i \in \mathbb{R}^{K \times M_i}$, where $M_i = H_i \times W_i$. To compute spatial weights for each class, we normalize the heatmap across the spatial dimension, yielding $\tilde{h}^{\text{flat}}_i$. The class-wise feature prototypes are then obtained as weighted sums of the feature vectors:
\begin{equation}
    \text{P}_i = \tilde{h}^{\text{flat}}_i \cdot {f^{\text{flat}}_i}^\top, \quad \text{P}_i \in \mathbb{R}^{K \times C_i}.
    \label{equ:prototype}
\end{equation}
Each row of $\text{P}_i$ represents the prototype feature vector for a specific class. For the $i$-th feature level, we obtain a set of prototypes $\{\text{P}^T_i, \text{P}^{S_1}_i, \dots, \text{P}^{S_N}_i\}$, corresponding to the target domain and each of the $N$ source domains. These prototypes are then $\ell_2$-normalized, and pairwise distances are computed as $1.0$ minus the cosine similarity between each pair of normalized vectors, resulting in a prototype graph for the $i$-th feature level. The final prototype graph $\mathcal{G}$ is obtained by averaging the graphs from the low-, middle-, and high-level features.

During inference, given the prototype graph $\mathcal{G} \in \mathbb{R}^{K \times (1+N) \times (1+N)}$ and the $N$ sets of predictions generated using different domain embeddings, we apply a weighted box fusion strategy to obtain the final fused predictions. Specifically, for each prediction consisting of a 3D bounding box $b$, confidence score $s$, and category label $l$, we first calculate its weight based on the prototype graph as follows:
\begin{equation}
    w = s / (1+\mathcal{G}_{l,1+n,1}),
\end{equation}
where $\mathcal{G}_{l, 1+n, 1}$ denotes the prototype distance between the $n$-th source domain and the target domain for category $l$. Next, for each category, the 3D Intersection-over-Union (IoU) is computed between all pairs of predictions belong to this category, and predictions are grouped into clusters if their IoU exceeds 0.1. For each resulting cluster $\mathcal{C}_i$, the fused bounding box $\hat{b}_i$ and confidence score $\hat{s}_i$ are computed as weighted averages of the predictions within the cluster as follows:
\begin{equation}
\begin{aligned}
\hat{b}_i &= \frac{\sum_{j \in \mathcal{C}_i} w_j b_j}{\sum_{j \in \mathcal{C}_i} w_j}, \quad
\hat{s}_i = \frac{\sum_{j \in \mathcal{C}_i} w_j s_j}{\sum_{j \in \mathcal{C}_i} w_j}
\end{aligned}
\end{equation}
Here, $\mathcal{C}_i$ denotes the set of indices corresponding to the predictions in the $i$-th cluster, and $j$ is an index within this set. To enable effective averaging of orientations, the rotation component of each bounding box is first converted to its sine and cosine representations before fusion. Note that the prototype graph only needs to be constructed once, offline, after training. During inference, the only additional computational cost comes from executing the fusion head $N-1$ extra times, which is relatively low compared to the cost of the feature backbones as shown in Tab. \ref{tab:components_time_consumption}.

\section{Experiments}

    \begin{table*}[!ht]
        \centering
        %\vspace{-0.3cm}
    	\caption{An ablation study on different pre-training domain adaptation (PTDA) strategies for single-source multi-modality domain adaptation from nuScenes to Waymo.}
    	\label{tab:pretrain_da_methods}
    	\small
    	\setlength{\tabcolsep}{2.5mm}    
            \renewcommand*{\arraystretch}{1.0}
    	\begin{tabular}{c|c|c|c|c|c|c|c|c|c}
    	\hline
            \multicolumn{4}{c|}{PTDA Strategy} &\multicolumn{3}{c|}{\makecell{mAP L1/mAP L2}} &\multicolumn{3}{c}{\makecell{mAPH L1/mAPH L2}}\\
            \hline
            \makecell{Coordinate\\ Shifting} &\makecell{Intensity\\Removal} &\makecell{Velocity\\Removal} &\makecell{Class\\Remapping} &Car &Pedestrian &Cyclist &Car &Pedestrian &Cyclist\\
            \hline
            $\times$ &$\times$ &$\times$ &$\times$ &25.8/22.0 &17.3/14.4 &2.1/2.0 &24.8/21.1 &9.4/7.8 &1.4/1.3 \\
            
            \checkmark &$\times$ &$\times$ &$\times$ &34.2/29.1 &16.0/13.2 &4.3/4.2 &32.8/28.0 &8.3/6.9 &2.7/2.6 \\

            \checkmark &\checkmark &$\times$ &$\times$ &44.0/37.5 &12.2/10.1 &5.2/5.0 &42.8/36.5 &6.8/5.6 &3.3/3.2 \\

            \checkmark &\checkmark &\checkmark &$\times$ &50.6/43.2 &17.9/14.8 &4.8/4.7 &49.3/42.0 &9.6/8.0 &3.3/3.2 \\    

            \checkmark &\checkmark &\checkmark &\checkmark &61.0/52.0 &55.5/46.0 &7.3/7.1 &59.1/50.4 &30.4/25.2 &4.8/4.6 \\      
    	\hline        
        \end{tabular}
    \end{table*}

    \begin{table*}[!ht]
        \centering
        %\vspace{-0.3cm}
    	\caption{Comparison of different methods on multi-source multi-modality domain adaptation from nuScenes, Lyft to Waymo. We implemented the pseudo label method in~\cite{inoue2018cross} and the disentangle learning method in~\cite{lin2021domain} for comparision. HSC-DC represents our proposed hierarchical spatially-conditioned domain classifiers, and PGW-MF denotes the prototype graph weighted multi-source fusion module.}
    	\label{tab:da_methods_msmm}
    	\small
    	\setlength{\tabcolsep}{2.5mm}    
            \renewcommand*{\arraystretch}{1.0}
    	\begin{tabular}{l|c|c|c|c|c|c}
    	\hline
            \multirow{2}{*}{Method} &\multicolumn{3}{c|}{mAP L1/mAP L2} &\multicolumn{3}{c}{mAPH L1/mAPH L2}\\
            
            \cline{2-7}
            &Car &Pedestrian &Cyclist &Car &Pedestrian &Cyclist\\
            \hline          

            Source Only (w/ PTDA) 
            &63.8/54.5 &53.2/44.1 &20.0/19.2 &62.4/53.2 &30.0/24.9 &15.3/14.7 \\        
            
            Pseudo label~\cite{inoue2018cross} 
            &67.4/57.5 &54.7/45.4 &20.7/19.9 &66.2/56.5 &32.6/27.0 &17.7/17.1\\
              
            Disentangle learning~\cite{lin2021domain}
            &66.9/57.1 &54.7/45.4 &20.4/19.6 &65.8/56.2 &32.2/26.7 &15.2/14.6\\

            HSC-DC (Our)
            &\textbf{68.9/58.8} &57.3/47.6 &23.0/22.1 &\textbf{67.8/57.9} &33.7/27.9 &18.8/18.1\\  
            
            HSC-DC + PGW-MF (Our)
            &68.2/58.2 &\textbf{63.5/52.7} &\textbf{28.0/27.0} &67.1/57.3 &\textbf{37.0/30.7} &\textbf{22.9/22.0}\\
            
            \cline{1-7}
            Oracle (Fully Supervised Target)
            &78.8/67.4 &83.7/69.7 &61.5/59.1 &77.7/66.4 &72.9/60.7 &59.6/57.3 \\  
            \hline
        \end{tabular}
    \end{table*}

\subsection{Setup Details}
\textbf{Datasets and Metrics}
We conduct experiments on three widely used multi-modal datasets that provide both LiDAR and camera data: Waymo~\cite{sun2020scalability}, nuScenes~\cite{caesar2020nuscenes}, and Lyft~\cite{lyft2019}. Since bounding boxes predicted in the UDA setting typically deviate more from the ground truth, we adopt relatively relaxed evaluation metrics. Specifically, when using Waymo as the target domain, we lower the default 3D IoU thresholds from 0.7, 0.5, and 0.5 to 0.5, 0.3, and 0.3 for Car, Pedestrian, and Cyclist, respectively, and report results on mAPL1/L2 as well as mAPHL1/L2. In the following experiments, unless otherwise specified, we use nuScenes and Lyft as the source domains and Waymo as the target domain. 

\textbf{Methods}
We adopt the widely used BEVFusion~\cite{liu2023bevfusion} as our baseline and implement two additional approaches — pseudo labeling~\cite{inoue2018cross} and disentangled learning~\cite{lin2021domain} — for comparison, given the absence of existing multi-source multi-modality methods. 

\textbf{Parameters}
We follow the default configuration of the BEVFusion implementation in mmdetection3d~\cite{mmdet3d2020}. The detection range is set to $[-54.0\text{m}, 54.0\text{m}]$ along the X and Y axes, and $[-5.0\text{m}, 3.0\text{m}]$ along the Z axis. The voxel size is $[0.075\text{m}, 0.075\text{m}, 0.2\text{m}]$ for the X, Y, and Z axes, respectively. All experiments are conducted using two RTX A6000 GPUs.

\textbf{Pre-training Domain Adaptation}
Simple preprocessing operations on the source domain prior to training have been shown to be highly effective for domain adaptation. For instance, in the early LiDAR-only UDA work SN~\cite{wang2020train}, source-domain bounding boxes were resized to better match the target-domain distribution. In the multi-modality setting, however, factors such as image resolution, point cloud coordinates, and intensity introduce additional complexity. As summarized in \tab~\ref{tab:pretrain_da_methods}, we evaluate four practical pre-training domain adaptation (PTDA) strategies — coordinate shifting, intensity removal, velocity removal, and class remapping — using the BEVFusion model trained on nuScenes and tested on Waymo. Coordinate shifting aligns the Z axis such that the ground plane corresponds to z=0. Intensity removal restricts input to the X, Y, and Z coordinates of the point cloud, thereby reducing the domain gap associated with intensity values. Velocity removal omits velocity prediction, allowing the model to focus exclusively on 3D bounding box estimation. Class remapping standardizes the label space across datasets by consolidating classes into Car, Pedestrian, Cyclist, and Others, and the model is trained to predict only Car, Pedestrian, and Cyclist. As shown in \tab~\ref{tab:pretrain_da_methods}, these PTDA strategies effectively reduce prominent domain gaps, resulting in more reasonable domain adaptation performance. The corresponding results serve as our comparison baseline.

    \begin{table}[!ht]
        \centering
        %\vspace{-0.3cm}
    	\caption{Ablation study on the effect of our proposed hierarchical spatially-conditioned domain classifiers (HSC-DC) strategy.}
    	\label{tab:ablation_dc}
    	\small
    	\setlength{\tabcolsep}{2.5mm}    
            \renewcommand*{\arraystretch}{1.0}
    	\begin{tabular}{l|c|c|c}
    	\hline
            \multirow{2}{*}{Method} &\multicolumn{3}{c}{mAP L1/mAP L2} \\
            
            \cline{2-4}
            &Car &Pedestrian &Cyclist\\
            \hline          

            Domain Classifier~\cite{hsu2020every} 
            &67.2/57.3 	&\textbf{60.7/50.3} 	&4.8/4.6  \\ 
            
            HSC-DC (Our)
            &\textbf{68.9/58.8} &57.3/47.6 &\textbf{23.0/22.1} \\  
            \hline
        \end{tabular}
    \end{table}

    \begin{table*}[!ht]
        \centering
        %\vspace{-0.3cm}
    	\caption{Ablation study on our proposed prototype graph weighted multi-source fusion (PGW-MF) strategy.``Source 1 Prediction" and ``Source 2 Prediction" denote the prediction results generated by the first and second fusion heads, respectively.}
    	\label{tab:ablation_mh}
    	\small
    	\setlength{\tabcolsep}{2.5mm}    
        \renewcommand*{\arraystretch}{1.0}
    	\begin{tabular}{l|c|c|c|c|c|c}
    	\hline
            \multirow{2}{*}{Method} &\multicolumn{3}{c|}{mAP L1/mAP L2} &\multicolumn{3}{c}{mAPH L1/mAPH L2}\\
            
            \cline{2-7}
            &Car &Pedestrian &Cyclist &Car &Pedestrian &Cyclist\\
            \hline          

            Source 1 Prediction 
            &67.4/57.5 &54.3/45.0 &23.5/22.6 &66.3/56.6 &32.0/26.5 &19.3/18.6 \\ 
            
            Source 2 Prediction 
            &61.0/52.1 &59.9/49.6 &20.6/19.8 &59.5/50.8 &34.0/28.2 &16.0/15.4\\

            PGW-MF (Our)
            &\textbf{68.2/58.2} &\textbf{63.5/52.7} &\textbf{28.0/27.0} &\textbf{67.1/57.3} &\textbf{37.0/30.7} &\textbf{22.9/22.0}\\ 
            \hline
        \end{tabular}
    \end{table*}

    \begin{figure*}
    	\centering
    	\footnotesize
    	\begin{tabular}{cc}
    		\includegraphics[height=0.25\linewidth]{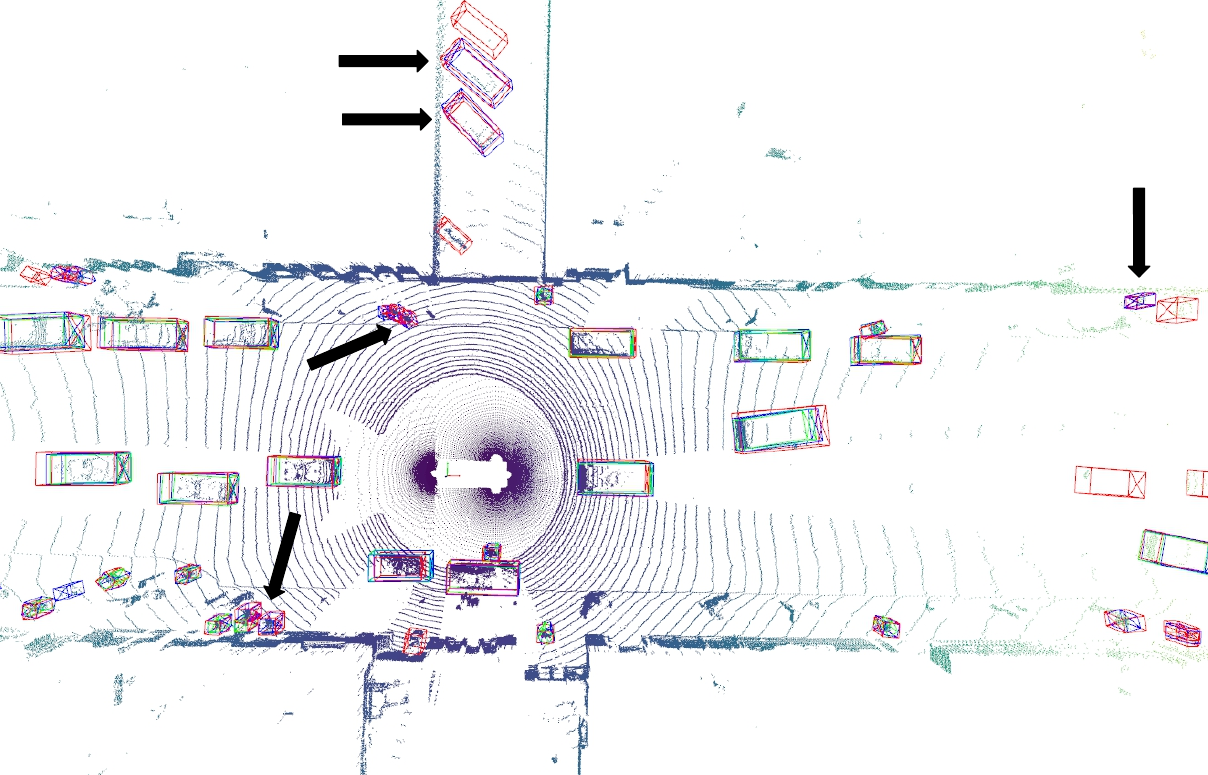} &\includegraphics[height=0.25\linewidth]{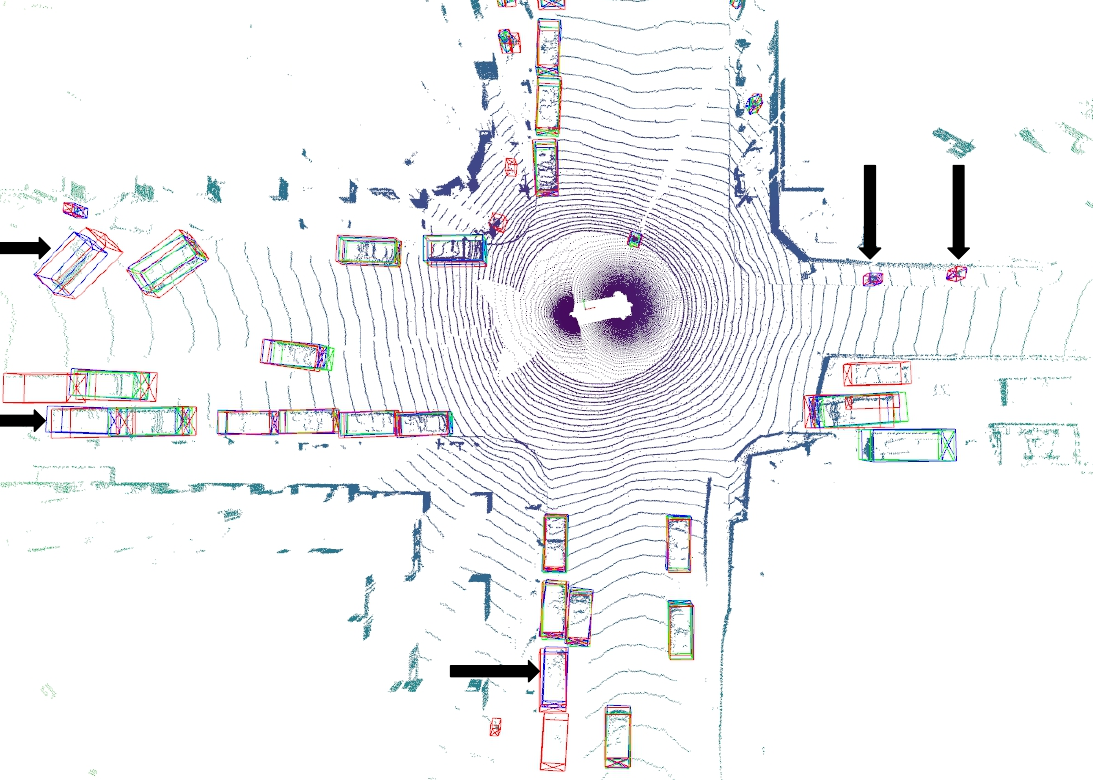}\\
            (a) &(b)\\
    	\end{tabular} 
    	\caption{Visual comparison between our method (HSC-DC + PGW-MF) and the Source Only baseline (w/ PTDA). The red, blue, and green bounding boxes denote the ground truth, our result, and the baseline result, respectively. Black arrows highlight our method's predictions miss-detected by the baseline.}
    	\label{fig:visual_results}
    \end{figure*} 
    
\subsection{Comparison}
As shown in \tab~\ref{tab:da_methods_msmm}, the Source Only (w/ PTDA) baseline exhibits the lowest performance across all object categories, indicating a pronounced domain gap when transferring from nuScenes and Lyft to Waymo even after applying pre-training domain adaptation strategies. Incorporating Pseudo Label or Disentangle Learning yields consistent improvements, demonstrating that leveraging target pseudo supervision or enforcing disentangled domain-invariant representations can partially mitigate domain shifts. Our proposed HSC-DC framework can improve performance across all metrics, confirming the advantage of hierarchical spatially-conditioned domain classifiers in aligning heterogeneous multi-source feature distributions. Integrating the PGW-MF module can further improve the performance, especially for challenging objects like Pedestrian and Cyclist, indicating that our prototype graph weighted fusion strategy can effectively capitalizes on complementary information across multiple source domains.

\subsection{Ablation Study}
\tab~\ref{tab:ablation_dc} shows the comparison between our proposed hierarchical spatially-conditioned domain classifiers and the traditional pixel-level domain classifier proposed in~\cite{hsu2020every}. As shown in the table, while the traditional domain classifier performs well on the Pedestrian category, it struggles significantly on Cyclist, leading to imbalanced performance across categories. By contrast, our HSC-DC achieves clear improvements on both Car and Cyclist, resulting in a more balanced overall performance. This indicates that by introducing spatial conditioning, the domain classifier can distribute attention more evenly across object categories, rather than being biased toward specific ones. In summary, incorporating spatial structure helps the model achieve more stable and reliable cross-domain feature alignment.

\tab~\ref{tab:ablation_mh} shows an ablation study assessing the effectiveness of the proposed PGW-MF multi-source fusion strategy. Across all object categories and metrics, PGW-MF consistently outperforms using either source model alone. For instance, PGW-MF improves Car mAP to 68.2/58.2 compared to 67.4/57.5 and 61.0/52.1 from the individual sources. The gains are more pronounced for Pedestrian and Cyclist, where PGW-MF effectively leverages complementary strengths of the two sources. Similar improvements in mAPH indicate that PGW-MF also enhances orientation estimation reliability. These results confirm that simply selecting one single source model is insufficient, and that PGW-MF provides a principled mechanism to integrate heterogeneous source predictions. By weighting contributions based on prototype-level feature distances, PGW-MF effectively emphasizes reliable predictions while suppressing misleading ones, leading to better performance.

\subsection{Visual Comparison}
\fig~\ref{fig:visual_results} presents a visual comparison between our method and the Source-Only baseline (w/ PTDA). As highlighted by the black arrows, our method produces more correct detections (blue bounding boxes) than the baseline (green bounding boxes). Notably, these additional true positives often occur in challenging scenarios, such as distant objects or cases with sparse point observations. This demonstrates that our method achieves more robust and reliable detection performance compared to the baseline.

\section{Conclusion}
In this work, we addressed the challenge of multi-source, multi-modality unsupervised domain adaptation for 3D object detection in autonomous driving. We introduced hierarchical spatially-conditioned domain classifiers to jointly align camera and LiDAR features across all source–target domain pairs. To further exploit cross-domain complementarity, we constructed a prototype distance graph to characterize domain relationships and guide fusion. Building on this, we proposed a prototype distance graph weighted multi-source fusion strategy that adaptively aggregates predictions from multiple source detectors. Extensive experiments on Waymo, nuScenes, and Lyft demonstrate that our framework delivers consistent improvements over existing state-of-the-art baselines.

\section*{Acknowledgment}
This work was supported in part by the Sony Research Award Program.
% \small
% \bibliographystyle{ieeenat_fullname}
% \clearpage
\bibliographystyle{IEEEtran}
\bibliography{main}

\end{document}